\documentclass[electronic]{vgtc}             

\graphicspath{{figures/}{pictures/}{images/}{./}} 

\usepackage{cite}
\usepackage{amsmath,amssymb,amsfonts}
\usepackage{algorithmic}
\usepackage{graphicx}
\usepackage{textcomp}
\usepackage{url}
\usepackage{booktabs}
\usepackage{times}                    
\usepackage{subcaption} 
\usepackage{enumitem}

\usepackage{tabu}                      
\usepackage{booktabs}                  
\usepackage{lipsum}                    
\usepackage{mwe}                       

\usepackage{mathptmx}                  

\onlineid{0}

\vgtccategory{Research}

\vgtcinsertpkg

\title{{\Large \textsc \mbox{---------PREPRINT---------}}\\
Improving Remote Sensing Classification using Topological Data Analysis and Convolutional Neural Networks}

\author{Aaryam Sharma\thanks{e-mail: a584shar@uwaterloo.ca}\\ %
        \scriptsize University of Waterloo \\
        }

\abstract{
    Topological data analysis (TDA) is a relatively new field that is gaining rapid adoption due to its robustness and ability to effectively describe complex datasets by quantifying geometric information. In imaging contexts, TDA typically models data as filtered cubical complexes from which we can extract discriminative features using persistence homology. Meanwhile, convolutional neural networks (CNNs) have been shown to be biased towards texture based local features. To address this limitation, we propose a TDA feature engineering pipeline and a simple method to integrate topological features with deep learning models on remote sensing classification. Our method improves the performance of a ResNet18 model on the EuroSAT dataset by 1.44\% achieving 99.33\% accuracy, which surpasses all previously reported single-model accuracies, including those with larger architectures, such as ResNet50 ($2\times$ larger) and XL Vision Transformers ($197\times$ larger). We additionally show that our method's accuracy is 1.82\% higher than our ResNet18 baseline on the RESISC45 dataset. To our knowledge, this is the first application of TDA features in satellite scene classification with deep learning. This demonstrates that TDA features can be integrated with deep learning models, even on datasets without explicit topological structures, thereby increasing the applicability of TDA. A clean implementation of our method will be made publicly available upon publication.
} 


\begin{document}


\firstsection{Introduction}

\maketitle

Topological data analysis (TDA) is a powerful tool that utilizes geometric information of a dataset to derive global descriptors of the data \cite{munch_users_2017,mardones2025tda}. TDA is robust to noise, well-suited to complex geometric data, and applicable across a wide range of domains \cite{zia_topological_2024}. Therefore, despite being a relatively new field, TDA has been used in areas such as the environmental sciences, medical imaging and even finance \cite{sena_topological_2021,garside_topological_2019,singh_topological_2023,chang_topological_2023,ko_novel_2023,zia_topological_2024}.

TDA offers several advantages in the environmental sciences including determinism, transparency, and the ability to work with relatively small datasets \cite{ver_hoef_primer_2023}. As a result, TDA has been applied to several tasks, such as retrieving climate zone patterns \cite{sena_topological_2021}, matching of aerosol depth patterns \cite{ofori-boateng_application_2021}, analyzing active wildfires \cite{kim_deciphering_2019}, and analyzing atmospheric river patterns \cite{muszynski_topological_2019}. TDA is intuitively suited to these datasets as they often have a geometric, graph based, or point cloud representation.

\begin{figure}[htbp]
    \centering
    \includegraphics[width=0.5\textwidth]{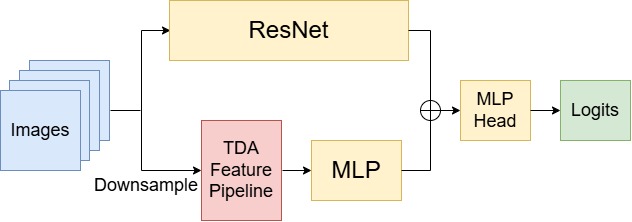} 
    \caption{Combining TDA and ResNet features via concatenation. Images are passed through a custom ResNet12 model or a ResNet18, and are also downsampled before passing through our TDA feature pipeline which are processed by a multi-layered perceptron (MLP) before being concatenated and then finally being passed through a 2-layered MLP to obtain logits.}
    \label{figure:model_diagram}
\end{figure}

Parallel to TDA, deep learning (DL) is another field that has seen an explosion in usage. Analyzing remote sensing data is a common application of machine learning to the environmental sciences. Remote sensing data is collected by satellites, drones, and other aerial surveillance technology often in the form of images. Remote sensing is being used to solve several pressing issues, such as climate change, increasing food security, and urban planning \cite{lu_lwganet_2025,yuan_deep_2020,mehmood_remote_2022,rolnick_tackling_2023}. A popular task in remote sensing is satellite scene classification, or land-use classification. In this task, satellite images are classified based on the type of terrain or land they depict - for example, industrial zones, herbaceous vegetation, or a sea/lake \cite{helber_eurosat_2019,helber_introducing_2018}. 

In recent years, the dominant models used for land-use classification are convolutional neural networks (CNNs) due to their strong performance on almost all image classification tasks. However, recent studies have shown that CNNs are biased towards high-frequency texture-based features \cite{geirhos_texture_bias, avidan_improving_2022}. Several satellite scenes might contain highly descriptive medium to low-frequency features, especially for similar landforms, which could appear as a limitation for traditional CNNs. On the other hand, topological data analysis is specially suited to handle global features that can provide the low-frequency features that CNNs often miss. 

Recent work suggests that fusing TDA based-features with some models could improve their performance on computer-vision tasks when applied to small datasets or in a domain where geometric information is important \cite{hajij_tda-net_2021,lima_image_2023}. However, there has been little work on combining TDA features with raw image features from CNNs on a typical medium-sized computer vision dataset.

In this paper, we present a method to combine features from TDA with techniques of deep learning and computer vision (\cref{figure:model_diagram}) to classify remote sensing data - a novel application domain for TDA to improve the performance of existing methods that use only TDA or only computer vision. The main contributions of this paper are:

\begin{itemize}[topsep=0pt,itemsep=0pt,parsep=0pt,partopsep=0pt]
    \item We design a robust feature engineering pipeline that combines multiple topological descriptors and uses filtrations based on image gradients and local entropy for the first time to our knowledge (refer to \cref{figure:tda_feature_pipeline}).
    \item We demonstrate the power of TDA features by proposing a simple lightweight model using both ResNet features with TDA features that can substantially increase both model convergence rate and model performance by 1.82\% on RESISC45 and 1.44\% on EuroSAT.
    \item We establish a new state-of-the-art (SOTA) classification accuracy of 99.33\% on the EuroSAT dataset for a single model (of any size) with a more than $2\times$ smaller model than ResNet50.
    \item We empirically demonstrate that EuroSAT contains globally discriminative features, while RESISC45 relies on local fine-grained features.
\end{itemize}


\section{Related Works}

TDA has been applied to computer vision tasks, where data is presented in an image or grid format with multiple bands. A common approach involves applying persistence homology to a filtered cubical complex (see Section 3: Mathematical Background) to generate features that can be used downstream for further analysis. Typically, these applications have relied on classical machine learning (ML) models such as support vector machines, linear or polynomial regression, random forests, gradient boosted trees, or other lightweight ML algorithms \cite{ver_hoef_primer_2023}.

To our knowledge, the first work bridging image applications and TDA was Garin et al. \cite{TDA_reading_lesson}, who explored the use of persistence homology on the MNIST dataset \cite{li_deng_mnist_2012}. Building on this, Lima et al. \cite{lima_image_2023} proposed a TDA feature pipeline for MNIST that when combined with a multi-layered perceptron (MLP) gave better performance than without these features, but was unable to substantially improve beyond the base CNN model.

A large amount of image based TDA work has also emerged out of the medical domain, such as analysis of MRI images \cite{singh_topological_2023} and diabetic retinopathy images \cite{garside_topological_2019}, typically also using classical ML models. Recently, a TDA-Net was proposed to analyze chest X-ray images for COVID-19 detection which combines DL with TDA, by converting a persistence diagram into a Betti curve as a vector \cite{hajij_tda-net_2021}. While this model can outperform baseline CNNs, the dataset used is very small (574 images). Furthermore, this work puts emphasis on developing a new architecture while only using a single persistence homology feature - Betti curves \cite{hajij_tda-net_2021}.

There has been significant research on building classification models for both the EuroSAT \cite{helber_eurosat_2019,helber_introducing_2018} and RESISC45 \cite{cheng_remote_2017} datasets. One of the first benchmark studies was conducted by Neumann et al. \cite{neumann_-domain_2019}. They achieved achieved an accuracy for ResNet50 on RESISC45 of 96.83\% and the SOTA accuracy for a ResNet50 model on EuroSAT of 99.20\%, by fine-tuning in-domain representations which were developed by training on a different full remote sensing dataset or ImageNet.

As larger models and ensembles have become more prevalent, many additional approaches have emerged. The current SOTA accuracy for EuroSAT is held by an ensemble of large deep models \cite{nanni_deep_2025}. Other approaches include multi-task pretraining \cite{wang_mtp_2024} where a foundational model is trained on a wide range of tasks before being evaluated on remote sensing tasks. Architectures optimized for efficiency, such as the lightweight group attention network LWGANet \cite{lu_lwganet_2025}, have also shown strong results. Another multitask system that was evaluated on EuroSAT and RESISC45 is presented in \cite{gesmundo_evolutionary_2022}, which, evolves its architecture and hyperparameters, and obtains the SOTA score for RESISC45 of 97\%.

To overcome the scarcity of labelled data in remote sensing, several works use self-supervised learning \cite{gomez_msmatch_2021}. Goyal et al. \cite{goyal_vision_2022} evaluated on EuroSAT and RESISC45 after a large amount of pretraining on uncurated data and scored high accuracies on both datasets. 

Despite these advances, many of these ideas are complex, use large models, use external data or in-domain pretraining. In contrast, our focus is instead on demonstrating the viability of TDA features themselves, using very small models and only training on the chosen datasets to show that TDA features can act as powerful data representations that can overcome the need for complex or expensive methodologies.

\section{Mathematical Background}
In this section, we describe the mathematical background required to explain the feature extraction pipeline in the next section. The following definitions are from \cite{mardones2025tda} and are presented to provide intuition about our pipeline. For a more rigorous look, we refer the reader to \cite{lima_image_2023,kaczynski_computational_2004,edelsbrunner_computational_2010}.


A \textbf{hypercube} in $\mathbb{R}^n$ is defined as a pair $(v, \sigma)$ where $v \in \mathbb{Z}^n$ is the anchor point, and $\sigma \in \{e, f\}^{\times n}$ is its extension descriptor. Here $\sigma[i] = e$ implies that the cube extends in the $i^{th}$ direction and is fixed otherwise. Its dimension is defined as $\#\{i: \sigma[i] = e\}$.

For each extendible coordinate $k$, the bottom face is the hypercube $(v, \sigma_k)$ where $\sigma_k[k] = f$ and $\forall i \ne k: \sigma_k[i] = \sigma[i]$, and the top face is the hypercube $(v_k, \sigma_k)$, where $v_k[k] = v[k] + 1$ and $\forall i \ne k: v_k[i] = v[i]$.

A \textbf{cubical complex} is a collection of hypercubes that is closed under taking faces. A \textbf{cubical subcomplex} of a cubical complex is a subset that is itself a cubical complex. A \textbf{top cube} is a cube that is not a subset of any other cube in the complex.

\begin{figure}[htbp]
    \centering
    \includegraphics[width=0.4\textwidth]{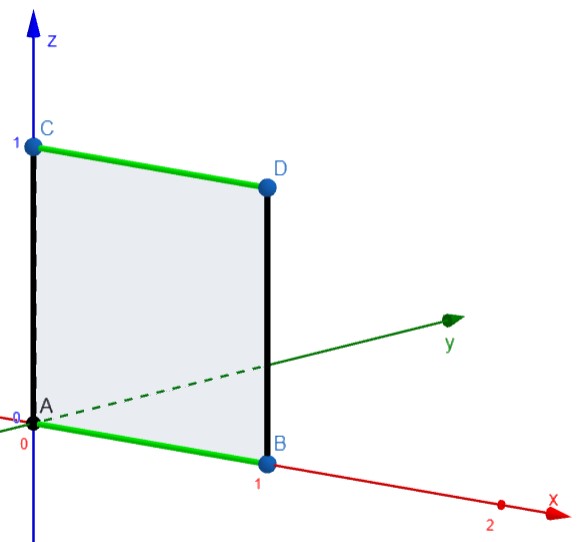}  
    \caption{ABDC is a cube $((0, 0, 0), (e, f, e))$. For the x-coordinate, the bottom face is AB and the top face is CD. For z-coordinate the faces are correspondingly AC and BD. ABDC, all faces and vertices, together form a cubical complex.}
    \label{figure:cubical_complex}
\end{figure}

A \textbf{filtered cubical complex} $C$ is a collection of cubical complices $\{C_t\}_{t\in \mathbb{R}}$ such that $\forall t \le s \in \mathbb{R}$, we have $C_t \subseteq C_s$. 

An $N_1 \times N_2 \times \dots \times N_D$ \textbf{grid complex} is a cubical complex with top dimensional cubes whose anchor points $(v_1, v_2, \dots, v_D)$, satisfy $v_i \in \{0, \dots, N_i - 1\}$, and extension descriptors are of the form $e^D$. We can represent a \textbf{grayscale digital $D$-dimensional image} as a $N_1 \times N_2 \times \dots \times N_D$ grid complex and a non-negative real function defined on its top cubes, e.g. pixel intensity. 



\subsection{Filtrations}

We can associate a filtered cubical complex with a grayscale image using several methods. The following is a list of the filtration methods used: 

\begin{enumerate}
    \item Binary threshold followed by height and radial filtrations.
    \item Grayscale filtration
    \item Local entropy filtration
    \item Image gradient filtration
\end{enumerate}

\begin{figure*}[htbp]
    \centering
    \includegraphics[width=0.9\textwidth]{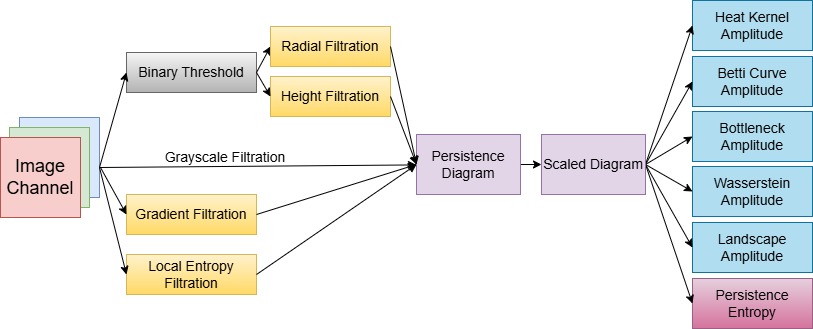}  
    \caption{The TDA pipeline in full. A specific image channel is selected, then passed through each of the filtrations. Before the radial or height filtration, images are first thresholded to make them binary. From these a persistence diagram is computed and then subsequently scaled. From the scaled diagram we compute the heat kernel, Betti curve, bottleneck, Wasserstein and landscape amplitudes, and also the persistence entropy as features.}
    \label{figure:tda_feature_pipeline}
\end{figure*}

\subsubsection{Binary Thresholding}

The distribution of color intensities was first examined for each class. Several classes have distinct mean intensities across color channels and some classes have non-overlapping error bars. This suggests that thresholding a channel at a set of intensities could yield class-distinct binary masks. Additionally, some thresholds highlight regions in images with especially high or low intensity across combinations of channels.

\begin{figure*}[htbp]
    \centering
    \includegraphics[width=\textwidth]{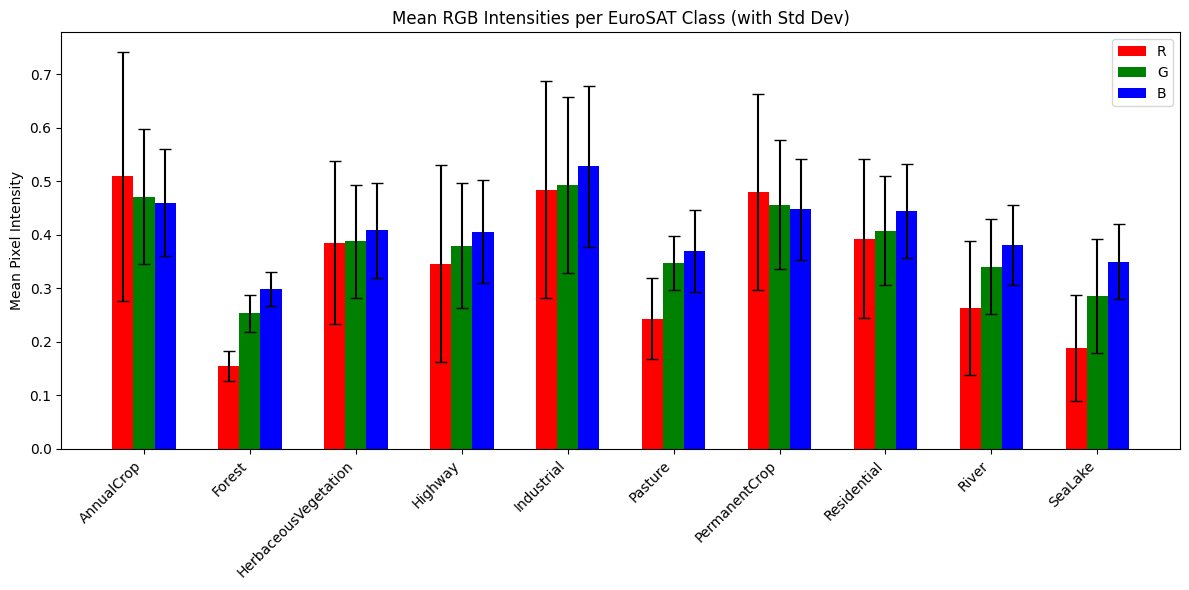}  
    \caption{Mean color intensity per class for each EuroSAT category. Some classes such as forests can be seen on average, to have very low Red channel intensity and low standard deviation, while Annual Crop, has a very high standard deviation but also a large mean. This supports the idea of applying selective thresholds to only retain on low red channel intensities and similarly retain only very high intensities. A similar pattern can be seen with the green and blue channels.}
    \label{figure:eurosat_intensity_barplot}
\end{figure*}

To obtain a filtration on these masked images, we use height and radial filtrations as outlined in \cite{lima_image_2023}. 

The height filtration on a binary image $\mathcal{B}$ for a selected direction is $v \in \mathbb{R}^2$ with norm 1 is defined as:
\begin{equation}
    \mathcal{H}_v(x) = 
    \begin{cases}
        \langle x, v \rangle & \text{if } \mathcal{B}(x) = 1 \\
        H_\infty & \text{otherwise}
    \end{cases}
    \label{height_filtration}
\end{equation}

where $H_\infty$ is the maximum value of $\langle x, v \rangle$ over all pixels $x$ in the image

The radial filtration on a binary image $\mathcal{B}$ for a selected center $c \in \mathbb{R}^2$ is defined as:
\begin{equation}
    \mathcal{R}_c(x) = 
    \begin{cases}
        \|x - c\|_2 & \text{if } \mathcal{B}(x) = 1 \\
        R_\infty & \text{otherwise}
    \end{cases}
    \label{radial_filtration}
\end{equation}
where $R_\infty$ is the maximum value of $\|x - c\|_2$ over all pixels $x$ in the image.

In our pipeline, height filtration is applied using the four cardinal directions: up, down, left, and right. As our images are rescaled to size $32 \times 32$, the chosen centers are $(8, 8)$, $(8, 24)$, $(24, 8)$, $(24, 24)$ for radial filtration. These correspond to the centers of the 4 quadrants of the rescaled image.

\subsubsection{Grayscale Filtration}

For grayscale filtration, no special processing is applied. An image channel $c$ is selected and is computed for image $I$ as
\begin{equation}
    G(x) = I[c][x] \label{grayscale_entropy}
\end{equation}
In this paper, the grayscale filtration is computed for each RGB channel.

\subsubsection{Local Entropy Filtration}

Certain regions of an image may have a high variation of pixel intensities. To capture this local complexity, local entropy is used for each available channel in kernels of size $3 \times 3$ and $5 \times 5$. The local entropy filtration for a kernel size $k$ on channel $C$ is defined as:
\begin{equation}
    L_k (x) = -\sum_{p \in \text{Hist}(K_k(C, x))} p \log_2 (p) 
    \label{local_entropy}
\end{equation}
Where $K_k(C, x)$ represents the set of pixels in channel $C$, in a $k \times k$ kernel around pixel $x$ and $\text{Hist}$ computes the relative frequency histogram of the channel intensities in the input pixel set (to get probabilities $p$).

\begin{figure*}[htbp]
    \centering
    \includegraphics[width=\textwidth]{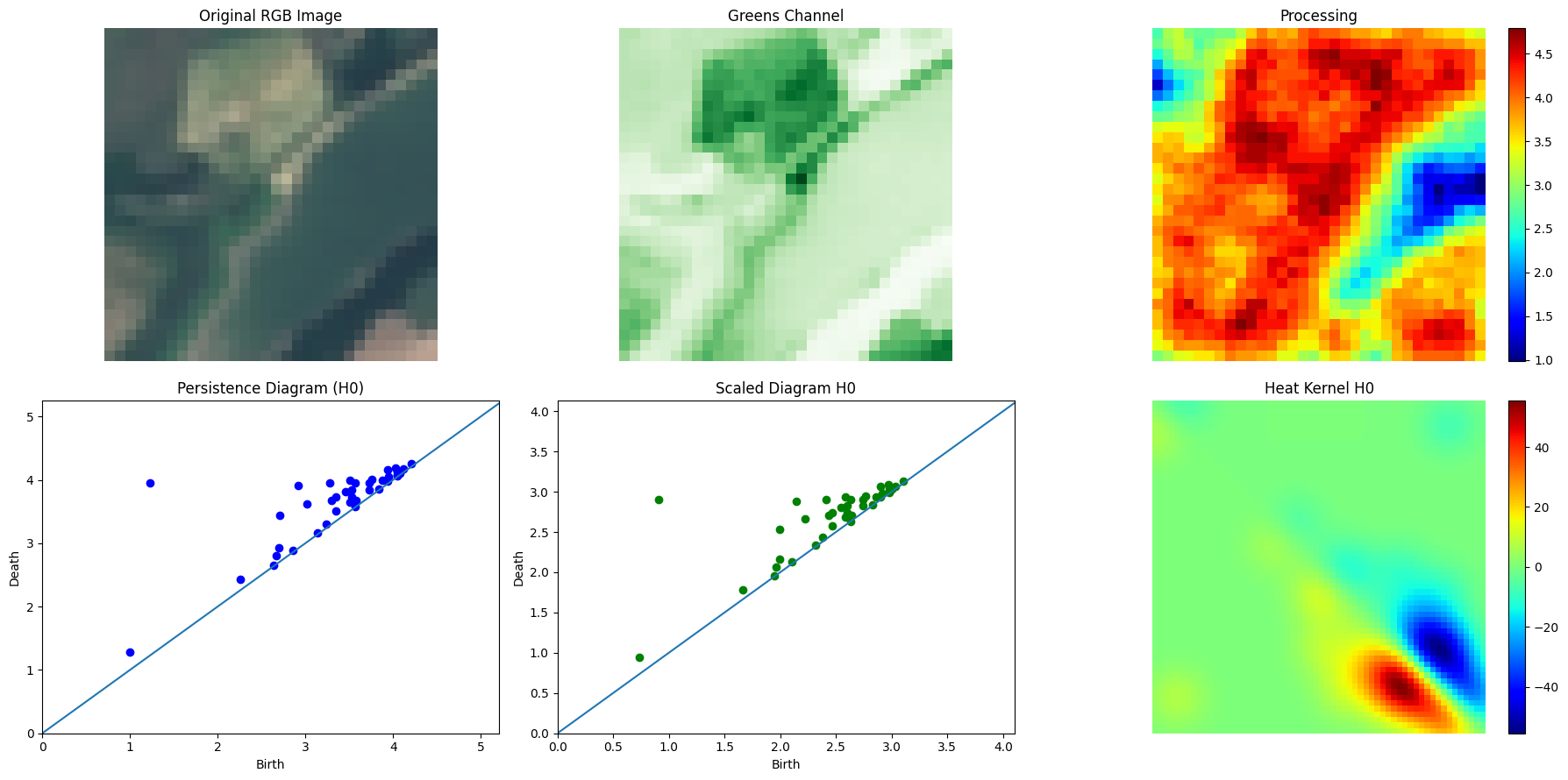}  
    \caption{An example of the pipeline of local entropy filtration is shown. Starting from the top left panel to the bottom right, we have (1) the original RGB Image, (2) its green channel selected, (3) local entropy applied to the green channel, (4) the persistence diagram of $H_0$, (5) a scaled version of (4) and finally its heat kernel in (6).  Across bodies of water, we can see low local entropy whereas, at points where different types of landmasses intersect, there is very high entropy}
    \label{figure:local_entropy_pipeline}
\end{figure*}

\subsubsection{Image Gradient Filtration}
Images of locations such as highways, residential areas and farms will typically have strong boundaries, each with distinct topological characteristics. For example, highways will usually have two straight boundaries, whereas a farmland will have multiple disconnected boundary components separating fields. These global, boundary-based properties can be captured effectively using image gradients by defining a filtration on a channel $C$ as:
\begin{eqnarray}
    IG_x(p) &=& \text{Conv}(G_x, C)[p] \\
    IG_y(p) &=& \text{Conv}(G_y, C)[p]
    \label{gradient_filtration}
\end{eqnarray}

Where $\text{Conv}(K, C)$ applies a convolution kernel $K$ to image $C$ (with appropriate padding to preserve dimensions). The kernels $G_x, G_y$ are the Sobel operators:

\begin{eqnarray}
    G_x &=&
\begin{bmatrix}
-1 & 0 & 1 \\
-2 & 0 & 2 \\
-1 & 0 & 1
\end{bmatrix}, \quad
G_y =
\begin{bmatrix}
-1 & -2 & -1 \\
0 & 0 & 0 \\
1 & 2 & 1
\end{bmatrix}
\end{eqnarray}

\subsection{Cubical Homology and Persistence Homology}

We can define vector spaces using a cubical complex $Q$, where, we treat each $d$-dimensional cube as a formal element and the corresponding vector space $Q_d$ as all linear combinations of these elements. These are formally known as \textbf{cubical chains}. 

We can then define the \textbf{boundary map} to be the linear map $\partial_d:Q_d \rightarrow Q_{d-1}$, defined on elementary cubes $c = (v, \sigma)$ as in \cref{eq:boundary_function}
\begin{equation}
    \partial_d (c) = \sum_{\substack{1 \le k \le n \\ \sigma[k] = e}} (-1)^{z(k)} \left(\text{top}_k(c) - \text{bottom}_k(c) \right)
    \label{eq:boundary_function}
\end{equation}
Where $z(k) = |\{i < k: \sigma[i] = e\}|$, $\text{top}_k(c)$ and $\text{bottom}_k(c)$ represent the top and bottom faces of $c$ respectively, corresponding to the $k^{\text{th}}$ dimension.

The quotient spaces $H_d(Q) = \ker(\partial_d) / \operatorname{im}(\partial_{d+1})$ are called the \textbf{degree $d$-homology} of the cubical complex $Q$. If we have a filtered cubical complex $(C_t)_{t\in \mathbb{R}}$, then we get the natural homomorphism between homology groups $f_d^{s,t} : H_d(C_s) \rightarrow H_d (C_t)$ induced by $C_s \subseteq C_t$ whenever $s \le t$. 

This gives us a \textbf{persistent vector space} which is a collection of vector spaces $V = \{V_t\}_{t\in \mathbb{R}}$ and a family of linear maps $V_{s,t} : V_s \rightarrow V_t$ for all $s \le t$ such that $ V_{t, r} \circ V_{s, t} = V_{s, r}$ for all $s \le t \le r$. In our case, we take $V_t = H_d(C_t)$ and the maps $V_{s,t}$ are the natural induced homomorphisms discussed above.

\subsection{Barcode and Metrics}

As persistent vector spaces are indexed by $t \in \mathbb{R}$, the notion of a \textbf{barcode} $bar(V)$ of a persistent vector space $V$ arises. This is defined as a multiset of intervals called bars. The number of bars containing any interval $[s, t]$ is equal to $\text{rank}(V_{s,t})$. The start of each interval (left-endpoint) is called its birth, and the end (right-endpoint) its death.

For our purposes, since we deal with discrete finite data, we may assume that our filtered complex has only a finite number of distinct complexes (i.e., only at finitely many values $t\in \mathbb{R}$ do we see different $C_t$). This results in finitely many bars in the so-obtained barcode using any of the aforementioned filtrations. These bars represent topological invariants and have several representations that can be used to extract meaningful features for analysis. 

Plotting the birth time of each bar on the $x$-axis and death time on $y$-axis, we obtain a scatter plot of points on or above the line $y = x$ (as birth time always occurs before death time). This scatterplot is known as the persistence diagram. An example is the 1st panel in the 2nd row of \cref{figure:local_entropy_pipeline}. 

In addition to visual representations, we can extract numerical descriptors from these diagrams for downstream tasks. The following are key representations and tools used in our analysis. Several of these are metrics from which we can extract the amplitude of a persistence diagram, each representing different information. Note that each method is applied to bars of a single homological dimension at a time, as our persistent vector space is constructed separately for each degree.

\subsubsection{Betti Curves and Betti Distance}
Define the Betti curve as the function $B: \mathbb{R} \rightarrow \mathbb{N}$ such that:
\begin{equation}
    B(x) = \left|\left\{i \mid [b_i, d_i) \in \operatorname{bar}(V),\ x \in [b_i, d_i)\right\}\right|
    \label{eq:betti_curve}
\end{equation}

$B(x)$ counts the number of bars that contain $x$. The amplitude of this curve is typically measured as the $L^p$ norm of this curve, where $p$ is treated as a hyperparameter.

\subsubsection{Bottleneck Distance}
The bottleneck distance between two diagrams $D_1$ and $D_2$ is the smallest $L^\infty$ norm, over all bijections $\gamma: D_1 \rightarrow D_2$ where these diagrams are extended to contain $y=x$,
\begin{equation}
    d_B(D_1, D_2) = \inf_{\gamma} \sup_{x \in D_1} \|x - \gamma(x)\|_\infty
    \label{eq:bottleneck_distance}
\end{equation}

The amplitude of a diagram $D_1$ is its distance from the empty diagram (i.e., a diagram consisting only of the diagonal $y=x$).

\subsubsection{Wasserstein Distance}
This is very similar to the bottleneck distance, but it is a generalization with hyperparameters $p, q$. The diagram is also extended to contain the line $y = x$ 
\begin{equation}
    d_W^{p,q}(D_1, D_2) = \inf_{\gamma} \left(\sum_{x \in D_1} \|x - \gamma(x)\|^p\right)^{\frac{1}{q}}
    \label{eq:wasserstein_distance}
\end{equation}

As before, the amplitude of a diagram $D_1$ is its Wasserstein distance to the empty diagram. We always set $q=p$ to ensure that our Wasserstein distance is a true metric.

\subsubsection{Landscape Distance}
Given a persistence diagram $D$, we can define a family of functions $\Lambda_i(x)$ as:
\begin{equation}
    \Lambda_i (x) = \max\left(0, \min(x - b_i, d_i - x)\right)
    \label{eq:landscape_basic}
\end{equation}
Each $\Lambda_i(x)$ is a triangular bump function representing the lifespan of the $i^{\text{th}}$ bar. The landscape functions $\lambda_k (x)$ are defined as the $k^{th}$ largest value amongst the $\{\Lambda_i(x)\}_{i\in \mathbb{N}}$. This is called the $k^{th}$ layer of our persistence landscape. The landscape distance between two landscapes $\lambda, \mu$ for a parameter $p \ge 1$:
\begin{equation}
    \|\lambda - \mu\|_p = \left(\sum_{i=1}^{\infty}\left(\int_{0}^{\infty} (\lambda_i (x) - \mu_i(x))^p dx\right)\right)^{\frac{1}{p}}
    \label{eq:landscape_distance}
\end{equation}
The amplitude of a diagram $D$ is thus its landscape distance to the empty diagram (whose landscape is all zeros).

\subsubsection{Heat Kernel and Amplitude}

We can apply a Gaussian kernel to each point in the persistence diagram and sum the resulting functions to get the heat kernel representation. Its amplitude is the $L^p$ norm of this resulting function in \cref{eq:heat_kernel} with hyperparameters $p, t$.
\begin{equation}
    h(x) = \sum_{[b, d) \in D} \frac{1}{4\pi t}\exp{\left(-\frac{\|x - (b, d)\|_2^2}{4t}\right)}
    \label{eq:heat_kernel}
\end{equation}

\subsubsection{Persistence Entropy}

While this is not a metric, a useful descriptor is the entropy of the persistence diagram $D$, defined to be \cref{eq:persistence_entropy}.

\begin{equation}
    E(D) = - \sum_{[b, d) \in D} \frac{d - b}{L_D} \log\left(\frac{d - b}{L_D}\right), \quad
    \text{where } L_D = \sum_{[b, d) \in D} (d - b)
    \label{eq:persistence_entropy}
\end{equation}

\section{Datasets and Evaluation Methodology}

\begin{figure*}[tb]
  \centering
  \begin{subfigure}[b]{0.2\linewidth}
    \includegraphics[width=\linewidth]{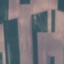}
    \caption{EuroSAT: Annual Crop}
  \end{subfigure}
  \hfill
  \begin{subfigure}[b]{0.2\linewidth}
    \includegraphics[width=\linewidth]{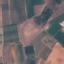}
    \caption{EuroSAT: Permanent Crop}
  \end{subfigure}
  \hfill
  \begin{subfigure}[b]{0.2\linewidth}
    \includegraphics[width=\linewidth]{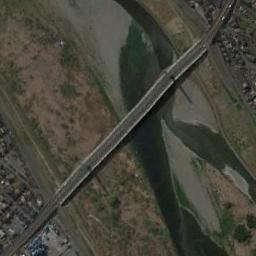}
    \caption{RESISC45: Bridge}
  \end{subfigure}
  \hfill
  \begin{subfigure}[b]{0.2\linewidth}
    \includegraphics[width=\linewidth]{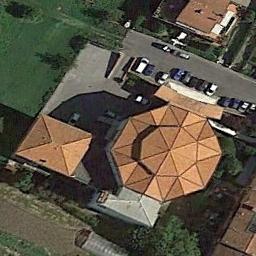}
    \caption{RESISC45: Church}
  \end{subfigure}
  \caption{Sample images from the EuroSAT and RESISC45 datasets. EuroSAT images capture land use patterns such as forest or industrial zones. RESISC45 includes object-centric classes like bridges and palaces, often centered in the image. We can also see that the Annual Crop and Permanent Crop categories look similar locally. On the other hand, RESISC45 have sharp local differences that are easy to spot. These details also get lost when we downsample the image before the TDA pipeline}
  \label{fig:dataset_examples}
\end{figure*}

We evaluate our method on two datasets in remote sensing classification: EuroSAT \cite{helber_eurosat_2019,helber_introducing_2018} and RESISC45 \cite{cheng_remote_2017}. These are some of the most commonly used remote sensing classification benchmarks. EuroSAT contains 27000 images across 10 landscape-based classes, with between 2000-3000 images per class. Several classes are very similar to an untrained eye. For example, annual crop and permanent crop both appear as farm-land with subtle differences. RESISC45 on the other hand contains 31500 images and 45 classes. This makes these datasets complementary in focus: classes in EuroSAT focus on the type of landscape such as forest, pasture or industrial areas, often characterized by distributed texture. In contrast, many RESISC45 classes focus on a specific central object such as a bridge, freeway, or even a palace (note that RESISC45 also includes classes such as forest and wetland, but many are specific central objects). See \cref{fig:dataset_examples}. 

These datasets are of the perfect size. The datasets are small enough that the topological features described earlier can be feasibly extracted for all images, while the datasets are several times larger than previous works combining CNNs and TDA. It is, however, known that many TDA algorithms computationally scale very rapidly with the amount of data \cite{zia_topological_2024}. Thus, for improving computational feasibility, the images are reduced to 32x32 before applying the topological data analysis pipeline.

Both datasets are split into a train, validation and test split in the ratio 60:20:20 as described in \cite{neumann_-domain_2019, wang_mtp_2024} although the exact same splits were not available for EuroSAT. This is the same evaluation split ratio as both these papers. These splits are fixed across all experiments to ensure consistency.

\section{Experiment Setup}

For both experiments, two backbones are used, a custom small ResNet model (ResNet12), and the other is a ResNet18. The best results of both models are presented in this paper. Images were rescaled for ResNet12 to $64 \times 64$, whereas for ResNet18 they were rescaled to $224 \times 224$. The ResNet12 model has $1.2M$ parameters, roughly $9\times$ smaller than ResNet18.

We follow a simple training procedure. Models were trained for only $50$ epochs each on cross-entropy loss, using Adam as the optimizer. A ReduceLROnPlateau scheduler was used to adjust learning rates dynamically. The learning rate for all ResNet12-based models was $3\times 10^{-4}$. The exact parameters for the ResNet18-based models can be found in our config file in our code. 

We extract TDA features using the giotto-tda library \cite{tauzin_giotto-tda_2021}. From our experimentation, it was found that in most images, there are very few dimension 1 features if any, so the processing is restricted to dimension 0 to avoid adding features that are almost entirely 0s. The TDA feature engineering pipeline can be found in \cref{figure:tda_feature_pipeline}. 

TDA features are then processed in all models using the same MLP component. This is a fully connected neural network with 4 hidden layers with ReLU activation, and two layers of dropout in the final two layers. We also used BatchNorm1d after each linear layer in this part of the model. These processed TDA features are concatenated with the features from the corresponding ResNet model before being passed through an output head that is a 2-layer MLP to obtain the logits. The logits are converted to probabilities using softmax. The TDA fusion model is presented in \cref{figure:model_diagram}.

For comparison, we also evaluate a TDA-only MLP model. This removes the CNN backbone and uses only the same 4-layer MLP and a 2-layer MLP output head, forming a 6-layer neural network, trained with the same procedure.

\section{Results}

The results of the experiments on both datasets are presented in \cref{table:EuroSATResult}, \cref{table:RESISCResult}. 

For EuroSAT, our TDA methodology with ResNet18 achieves 99.33\% accuracy, establishing a new state-of-the-art score for a single model (regardless of size). This is a 1.44\% improvement over our baseline ResNet18 model, and also better than the more than twice as large ResNet50 model that scores 99.2\% accuracy and even compared to much larger models such as InternImage-XL (which is $197\times$ larger). Furthermore, our ResNet12 model has 1.25M parameters, and our TDA module has 450K parameters. Our combined ResNet12 TDA model has 1.7M parameters which is $6.9\times$ smaller than ResNet18, yet, comes within 2\% of a plain ResNet18 model. 

This shows that with the choice of topological features, it is possible to greatly improve the performance of a significantly smaller model over previous SOTA approaches. This is with a simple training strategy compared to top-performing models, where we use a plain training loop with only a ReduceLROnPlateau scheduler.

For RESISC45, we are able to achieve 93.22\% accuracy, an improvement of 1.82\% in accuracy over the baseline. In both cases, topological data analysis results in more generalizable features than those learned purely from image data. Not only does the training accuracy improve, but the model's overfitting also decreases (smaller training vs test accuracy gaps \cref{table:EuroSATResult,table:RESISCResult}).

This also aligns with prior research \cite{geirhos_texture_bias, avidan_improving_2022} that shows that CNNs tend to focus on high-frequency texture based features. In our modelling scenario, satellite scenes contain strong features that may not appear on a local level (for example, Annual Crop and Permanent Crop might not seem very different to an untrained observer at a local level). Differences that smaller CNNs are not able to capture effectively. However, using TDA to obtain the same features more precisely, we can gain all the benefits of a smaller model, such as fast convergence and a simpler optimization, while significantly improving its representational capacity. 

This provides an intuitive explanation for why our RESISC45 model cannot match the accuracy of larger models as in EuroSAT. In RESISC45, as many classes focus on objects and structures, local features are more discriminative of the correct class. However, in EuroSAT, other than a few classes (such as highway, river and Sea/Lake) no such central object is typically present. Thus, topological features are more beneficial for EuroSAT than RESISC45 in allowing it to bridge the gap to larger models.

Further evidence is seen when we compare the MLPs trained only on the generated topological features. We find that for EuroSAT, this MLP alone reaches 92.39\% test accuracy, despite being a small 6-layer multi-layer perceptron. This, in turn, boosts the performance of both our ResNet models. On the other hand, for RESISC45, the same MLP only has 37.19\% accuracy. In fact, with the smaller ResNet12, TDA features appear to act as noise—reducing accuracy by nearly 5\%. This further suggests that the TDA features on RESISC45 are not as discriminative and likely become noise for the output head when our small ResNet12 is unable to capture deep features effectively.

This supports the view that EuroSAT contains more geometric, global features that TDA can model effectively, whereas the smaller, localized structures in RESISC45 can best be captured with larger CNNs with a small boost coming from TDA features.

\begin{table}[htbp]
\caption{EuroSAT Accuracy}
\begin{center}
\begin{tabular}{@{}llll@{}}
\toprule
Model                    & Train  & Validation  & Test  \\ \midrule
MSMatch \cite{gomez_msmatch_2021}                  & -                    & -                    & 98.65               \\
SEER (RG-10B) \cite{goyal_vision_2022}            & -                    & -                    & 97.5                \\
$\mu$2Net \cite{gesmundo_evolutionary_2022}                & -                    & -                    & 99.2       \\
Deep Ensembling \cite{nanni_deep_2025}          & -                    & -                    & \textbf{99.41}      \\
InternImage-XL (MTP+IMP) \cite{wang_mtp_2024} & -                    & -                    & 99.24               \\
ResNet50 (In Domain) \cite{neumann_-domain_2019}     & -                    & -                    & 99.2
\\ \midrule
ResNet12 (Our)           & 97.36                & 95.80                & 96.11               \\
ResNet18 (Our)           & 99.53                & 98.06                & 97.89               \\
TDA MLP                  & 97.56                & 93.26                & 92.39               \\
ResNet12 + TDA           & 99.63                & 97.72                & 98.07               \\
ResNet18 + TDA           & \underline{\textbf{99.98}} & \underline {\textbf{99.35}} & \underline {\textbf{99.33}} \\ \bottomrule
\label{table:EuroSATResult}
\end{tabular}
\end{center}
\end{table}

\begin{table}[htbp]
\caption{RESISC45 Accuracy}
\begin{center}
\begin{tabular}{@{}llll@{}}
    \toprule
    Model                    & Train  & Validation & Test \\ \midrule
    SEER (RG-10B) \cite{goyal_vision_2022}            & -     & -          & 95.61         \\
    $\mu$2Net \cite{gesmundo_evolutionary_2022}                & -     & -          & \textbf{97.0} \\
    InternImage-XL (MTP+IMP) \cite{wang_mtp_2024} & -     & -          & 96.27         \\
    ResNet50 (In Domain) \cite{neumann_-domain_2019}     & -     & -          & 96.8          \\  \midrule
    ResNet12 (Our)           & 91.80  & 82.73 & 81.41 \\
ResNet18 (Our)           & 97.53 & 92.10  & 91.40 \\
TDA MLP                  & 47.62 & 38.27 & 37.19 \\
ResNet12 + TDA           & 90.83 & 78.48 & 76.70 \\
ResNet18 + TDA           & \underline {\textbf{99.62}} & \underline {\textbf{94.02}} & \underline {\textbf{93.22}} \\ \bottomrule
\label{table:RESISCResult}
\end{tabular}
\end{center}
\end{table}

\begin{figure}[htbp]
    \centering
    \includegraphics[width=0.95\linewidth]{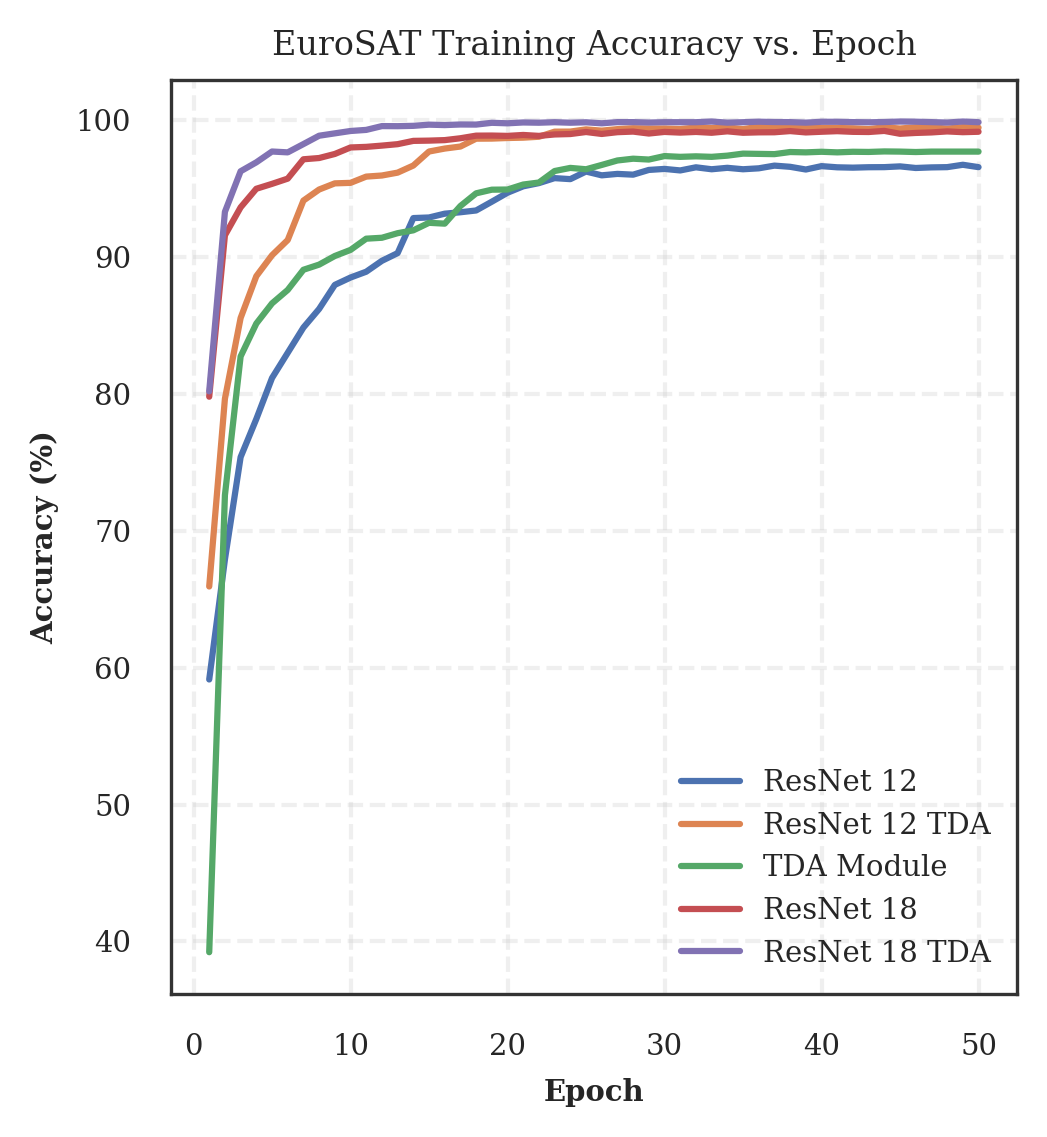}
    \caption{Accumulated training accuracy as the training epoch increases on EuroSAT. We see a significantly faster convergence using topological features for all models, including the TDA module alone.}
    \label{figure:train_accuracy}
\end{figure}

\begin{figure}[htbp]
    \centering
    \includegraphics[width=0.95\linewidth]{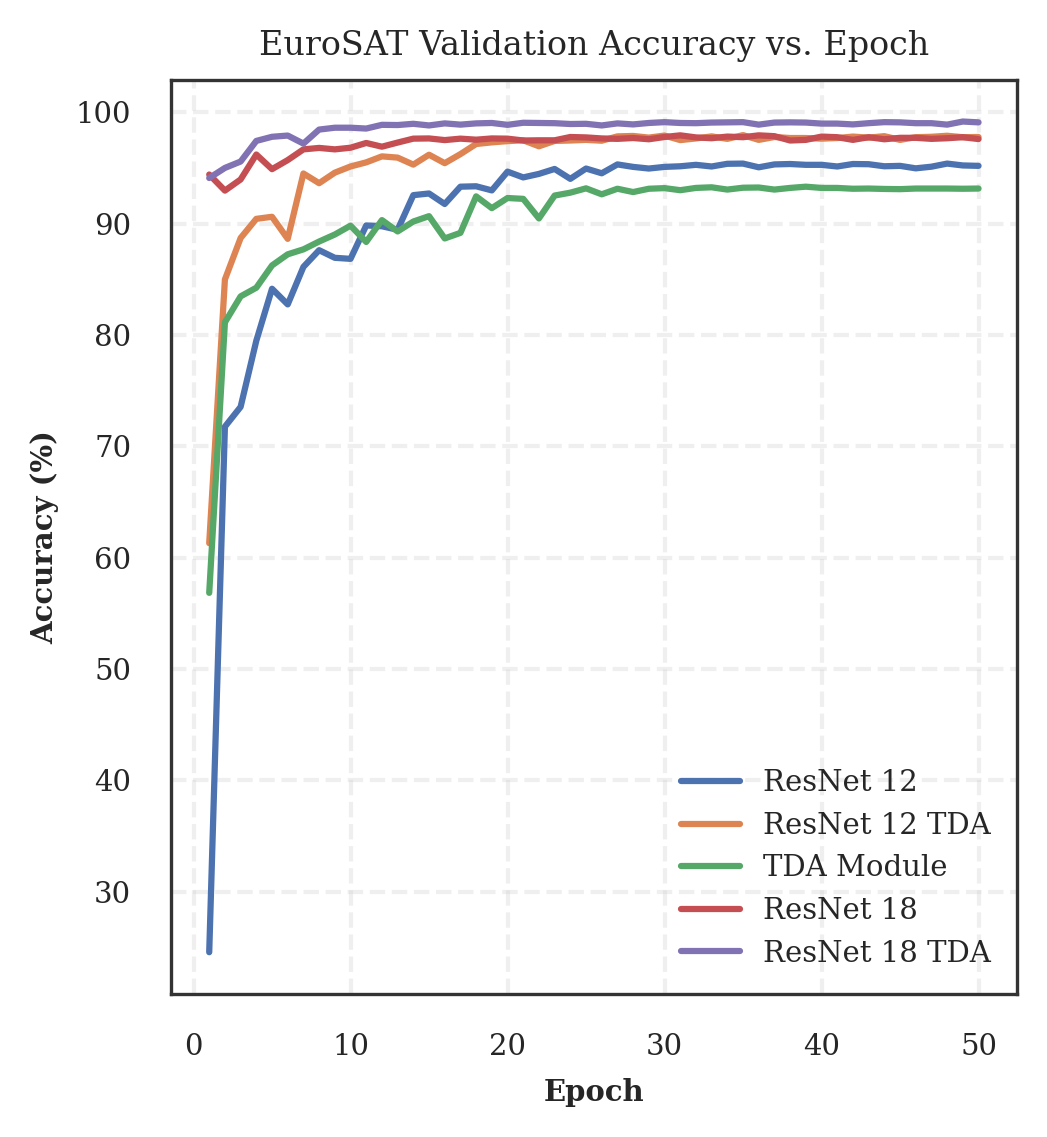}
    \caption{Validation accuracy as the number of epochs increases for the ResNet18 models on EuroSAT. We see a similar trend to the training accuracy, but a slightly noisier convergence. Nevertheless, using topological features speeds up convergence and allows the model to reach a higher accuracy. Note that the models start at a higher validation accuracy as these are computed post epoch, whereas accumulated training accuracy are collected from the start of each epoch, which has a lesser trained model than at the end.}
    \label{figure:validation_accuracy}
\end{figure}

As shown in \cref{figure:train_accuracy} and \cref{figure:validation_accuracy}, we can see that for EuroSAT, training both ResNets have faster convergence using topological features than without. This is also likely because the features themselves are quite descriptive (as even a 6-layer MLP learns very effectively), so our ResNet models learn much quicker.

\section{Conclusion}

In this paper, we have demonstrated that TDA can be effectively integrated with deep learning techniques using minimal architectural modifications. This enables small models to reach state-of-the-art performance on image classification tasks in remote sensing while also accelerating model convergence.

Our results demonstrate that TDA, when applied through a combination of multiple image filtrations, can extend beyond geometric datasets and contribute meaningfully to a broader class of visual classification problems by providing powerful global feature representations.

\section{Further Directions}

One promising extension of this work is exploring more deeply integrated architectures such as in TDA-NET \cite{hajij_tda-net_2021}, which may offer stronger feature fusion than simple concatenation. 

Another direction would be to expand this work to larger remote sensing classification datasets, or apply this approach in other domains where global features are strongly prevalent. One could also propose more filtrations and metrics to add diversity to the features.

We also downsampled our images greatly to generate TDA features for computational feasibility. This loses a lot of granularity in our TDA features. This likely made TDA features on RESISC45 far less effective than they could be. Thus, another direction could be to try this analysis on full images with sufficient resources.

While we were able to decrease the model sizes significantly, we also increased the inference time greatly because computing TDA features is very computationally intensive. This limitation suggests another direction in approximation techniques that could speed up feature processing pipelines greatly. 

\acknowledgments{
I gratefully acknowledge Professor Medina-Mardones for his encouragement and direction throughout this project.
}

\bibliographystyle{abbrv-doi}

\bibliography{template}

\begin{thebibliography}{10}

\bibitem{avidan_improving_2022}
J.~Bai, L.~Yuan, S.-T. Xia, S.~Yan, Z.~Li, and W.~Liu.
\newblock Improving {Vision} {Transformers} by {Revisiting} {High}-{Frequency}
  {Components}.
\newblock In S.~Avidan, G.~Brostow, M.~Ciss{\'e}, G.~M. Farinella, and
  T.~Hassner, eds., {\em Computer {Vision} - {ECCV} 2022}, vol. 13684, pp.
  1--18. Springer Nature Switzerland, Cham, Nov. 2022.
\newblock Series Title: Lecture Notes in Computer Science. doi: {{%
10\hspace{.1pt}\discretionary{.}{%
}{.}\hspace{.4pt}1007\discretionary{/}{%
}{/}978\discretionary{%
}{-}{-}3\discretionary{%
}{-}{-}031\discretionary{%
}{-}{-}20053\discretionary{%
}{-}{-}3\_1}}


\bibitem{chang_topological_2023}
C.~Chang and H.~Lin.
\newblock A topological based feature extraction method for the stock market.
\newblock {\em Data Science in Finance and Economics}, 3(3):208--229, July
  2023. doi: {{%
10\hspace{.1pt}\discretionary{.}{%
}{.}\hspace{.4pt}3934\discretionary{/}{%
}{/}DSFE\hspace{.1pt}\discretionary{.}{%
}{.}\hspace{.4pt}2023013}}


\bibitem{cheng_remote_2017}
G.~Cheng, J.~Han, and X.~Lu.
\newblock Remote {Sensing} {Image} {Scene} {Classification}: {Benchmark} and
  {State} of the {Art}.
\newblock {\em Proceedings of the IEEE}, 105(10):1865--1883, Oct. 2017. doi:
  {{%
10\hspace{.1pt}\discretionary{.}{%
}{.}\hspace{.4pt}1109\discretionary{/}{%
}{/}JPROC\hspace{.1pt}\discretionary{.}{%
}{.}\hspace{.4pt}2017\hspace{.1pt}\discretionary{.}{%
}{.}\hspace{.4pt}2675998}}


\bibitem{edelsbrunner_computational_2010}
H.~Edelsbrunner and J.~Harer.
\newblock {\em Computational topology: an introduction}.
\newblock American Mathematical Society, Providence, R.I, 2010.
\newblock OCLC: ocn427757156.

\bibitem{TDA_reading_lesson}
A.~Garin and G.~Tauzin.
\newblock A topological "reading" lesson: Classification of mnist using tda.
\newblock In {\em 2019 18th IEEE International Conference On Machine Learning
  And Applications (ICMLA)}, pp. 1551--1556, Dec. 2019. doi: {{%
10\hspace{.1pt}\discretionary{.}{%
}{.}\hspace{.4pt}1109\discretionary{/}{%
}{/}ICMLA\hspace{.1pt}\discretionary{.}{%
}{.}\hspace{.4pt}2019\hspace{.1pt}\discretionary{.}{%
}{.}\hspace{.4pt}00256}}


\bibitem{garside_topological_2019}
K.~Garside, R.~Henderson, I.~Makarenko, and C.~Masoller.
\newblock Topological data analysis of high resolution diabetic retinopathy
  images.
\newblock {\em PLOS ONE}, 14(5):e0217413, May 2019. doi: {{%
10\hspace{.1pt}\discretionary{.}{%
}{.}\hspace{.4pt}1371\discretionary{/}{%
}{/}journal\hspace{.1pt}\discretionary{.}{%
}{.}\hspace{.4pt}pone\hspace{.1pt}\discretionary{.}{%
}{.}\hspace{.4pt}0217413}}


\bibitem{geirhos_texture_bias}
R.~Geirhos, P.~Rubisch, C.~Michaelis, M.~Bethge, F.~A. Wichmann, and
  W.~Brendel.
\newblock Imagenet-trained cnns are biased towards texture; increasing shape
  bias improves accuracy and robustness.
\newblock In {\em 7th International Conference on Learning Representations,
  {ICLR} 2019, New Orleans, LA, USA, May 6-9, 2019}. OpenReview.net, May 2019.

\bibitem{gesmundo_evolutionary_2022}
A.~Gesmundo and J.~Dean.
\newblock An {Evolutionary} {Approach} to {Dynamic} {Introduction} of {Tasks}
  in {Large}-scale {Multitask} {Learning} {Systems}, Nov. 2022.
\newblock arXiv:2205.12755 [cs]. doi: {{%
10\hspace{.1pt}\discretionary{.}{%
}{.}\hspace{.4pt}48550\discretionary{/}{%
}{/}arXiv\hspace{.1pt}\discretionary{.}{%
}{.}\hspace{.4pt}2205\hspace{.1pt}\discretionary{.}{%
}{.}\hspace{.4pt}12755}}


\bibitem{gomez_msmatch_2021}
P.~Gomez and G.~Meoni.
\newblock {MSMatch}: {Semisupervised} {Multispectral} {Scene} {Classification}
  {With} {Few} {Labels}.
\newblock {\em IEEE Journal of Selected Topics in Applied Earth Observations
  and Remote Sensing}, 14:11643--11654, Nov. 2021. doi: {{%
10\hspace{.1pt}\discretionary{.}{%
}{.}\hspace{.4pt}1109\discretionary{/}{%
}{/}JSTARS\hspace{.1pt}\discretionary{.}{%
}{.}\hspace{.4pt}2021\hspace{.1pt}\discretionary{.}{%
}{.}\hspace{.4pt}3126082}}


\bibitem{goyal_vision_2022}
P.~Goyal, Q.~Duval, I.~Seessel, M.~Caron, I.~Misra, L.~Sagun, A.~Joulin, and
  P.~Bojanowski.
\newblock Vision {Models} {Are} {More} {Robust} {And} {Fair} {When}
  {Pretrained} {On} {Uncurated} {Images} {Without} {Supervision}, Feb. 2022.
\newblock arXiv:2202.08360 [cs]. doi: {{%
10\hspace{.1pt}\discretionary{.}{%
}{.}\hspace{.4pt}48550\discretionary{/}{%
}{/}arXiv\hspace{.1pt}\discretionary{.}{%
}{.}\hspace{.4pt}2202\hspace{.1pt}\discretionary{.}{%
}{.}\hspace{.4pt}08360}}


\bibitem{hajij_tda-net_2021}
M.~Hajij, G.~Zamzmi, and F.~Batayneh.
\newblock {TDA}-{Net}: {Fusion} of {Persistent} {Homology} and {Deep}
  {Learning} {Features} for {COVID}-19 {Detection} {From} {Chest} {X}-{Ray}
  {Images}.
\newblock In {\em 2021 43rd {Annual} {International} {Conference} of the {IEEE}
  {Engineering} in {Medicine} \& {Biology} {Society} ({EMBC})}, pp. 4115--4119.
  IEEE, Mexico, Nov. 2021. doi: {{%
10\hspace{.1pt}\discretionary{.}{%
}{.}\hspace{.4pt}1109\discretionary{/}{%
}{/}EMBC46164\hspace{.1pt}\discretionary{.}{%
}{.}\hspace{.4pt}2021\hspace{.1pt}\discretionary{.}{%
}{.}\hspace{.4pt}9629828}}


\bibitem{helber_introducing_2018}
P.~Helber, B.~Bischke, A.~Dengel, and D.~Borth.
\newblock Introducing {Eurosat}: {A} {Novel} {Dataset} and {Deep} {Learning}
  {Benchmark} for {Land} {Use} and {Land} {Cover} {Classification}.
\newblock In {\em {IGARSS} 2018 - 2018 {IEEE} {International} {Geoscience} and
  {Remote} {Sensing} {Symposium}}, pp. 204--207. IEEE, Valencia, July 2018.
  doi: {{%
10\hspace{.1pt}\discretionary{.}{%
}{.}\hspace{.4pt}1109\discretionary{/}{%
}{/}IGARSS\hspace{.1pt}\discretionary{.}{%
}{.}\hspace{.4pt}2018\hspace{.1pt}\discretionary{.}{%
}{.}\hspace{.4pt}8519248}}


\bibitem{helber_eurosat_2019}
P.~Helber, B.~Bischke, A.~Dengel, and D.~Borth.
\newblock {EuroSAT}: {A} {Novel} {Dataset} and {Deep} {Learning} {Benchmark}
  for {Land} {Use} and {Land} {Cover} {Classification}.
\newblock {\em IEEE Journal of Selected Topics in Applied Earth Observations
  and Remote Sensing}, 12(7):2217--2226, July 2019. doi: {{%
10\hspace{.1pt}\discretionary{.}{%
}{.}\hspace{.4pt}1109\discretionary{/}{%
}{/}JSTARS\hspace{.1pt}\discretionary{.}{%
}{.}\hspace{.4pt}2019\hspace{.1pt}\discretionary{.}{%
}{.}\hspace{.4pt}2918242}}


\bibitem{kaczynski_computational_2004}
T.~Kaczynski, K.~Mischaikow, and M.~Mrozek.
\newblock {\em Computational {Homology}}, vol. 157 of {\em Applied
  {Mathematical} {Sciences}}.
\newblock Springer New York, New York, NY, 2004. doi: {{%
10\hspace{.1pt}\discretionary{.}{%
}{.}\hspace{.4pt}1007\discretionary{/}{%
}{/}b97315}}


\bibitem{kim_deciphering_2019}
H.~Kim and C.~Vogel.
\newblock Deciphering {Active} {Wildfires} in the {Southwestern} {USA} {Using}
  {Topological} {Data} {Analysis}.
\newblock {\em Climate}, 7(12):135, Nov. 2019. doi: {{%
10\hspace{.1pt}\discretionary{.}{%
}{.}\hspace{.4pt}3390\discretionary{/}{%
}{/}cli7120135}}


\bibitem{ko_novel_2023}
S.~Ko and D.~Koo.
\newblock A novel approach for wafer defect pattern classification based on
  topological data analysis.
\newblock {\em Expert Systems with Applications}, 231:120765, Nov. 2023. doi:
  {{%
10\hspace{.1pt}\discretionary{.}{%
}{.}\hspace{.4pt}1016\discretionary{/}{%
}{/}j\hspace{.1pt}\discretionary{.}{%
}{.}\hspace{.4pt}eswa\hspace{.1pt}\discretionary{.}{%
}{.}\hspace{.4pt}2023\hspace{.1pt}\discretionary{.}{%
}{.}\hspace{.4pt}120765}}


\bibitem{li_deng_mnist_2012}
{Li Deng}.
\newblock The {MNIST} {Database} of {Handwritten} {Digit} {Images} for
  {Machine} {Learning} {Research} [{Best} of the {Web}].
\newblock {\em IEEE Signal Processing Magazine}, 29(6):141--142, Nov. 2012.
  doi: {{%
10\hspace{.1pt}\discretionary{.}{%
}{.}\hspace{.4pt}1109\discretionary{/}{%
}{/}MSP\hspace{.1pt}\discretionary{.}{%
}{.}\hspace{.4pt}2012\hspace{.1pt}\discretionary{.}{%
}{.}\hspace{.4pt}2211477}}


\bibitem{lima_image_2023}
M.~D.~P. Lima, G.~A. Giraldi, and G.~F.~M. Junior.
\newblock Image {Classification} using {Combination} of {Topological}
  {Features} and {Neural} {Networks}, Nov. 2023.
\newblock arXiv:2311.06375 [cs]. doi: {{%
10\hspace{.1pt}\discretionary{.}{%
}{.}\hspace{.4pt}48550\discretionary{/}{%
}{/}arXiv\hspace{.1pt}\discretionary{.}{%
}{.}\hspace{.4pt}2311\hspace{.1pt}\discretionary{.}{%
}{.}\hspace{.4pt}06375}}


\bibitem{lu_lwganet_2025}
W.~Lu, S.-B. Chen, C.~H.~Q. Ding, J.~Tang, and B.~Luo.
\newblock {LWGANet}: {A} {Lightweight} {Group} {Attention} {Backbone} for
  {Remote} {Sensing} {Visual} {Tasks}, Jan. 2025.
\newblock arXiv:2501.10040 [cs]. doi: {{%
10\hspace{.1pt}\discretionary{.}{%
}{.}\hspace{.4pt}48550\discretionary{/}{%
}{/}arXiv\hspace{.1pt}\discretionary{.}{%
}{.}\hspace{.4pt}2501\hspace{.1pt}\discretionary{.}{%
}{.}\hspace{.4pt}10040}}


\bibitem{mardones2025tda}
A.~M. Medina-Mardones.
\newblock Introduction to topological data analysis, 2025.
\newblock Course, Fields Institute, Jan.--Apr. 2025.

\bibitem{mehmood_remote_2022}
M.~Mehmood, A.~Shahzad, B.~Zafar, A.~Shabbir, and N.~Ali.
\newblock Remote {Sensing} {Image} {Classification}: {A} {Comprehensive}
  {Review} and {Applications}.
\newblock {\em Mathematical Problems in Engineering}, 2022:1--24, Aug. 2022.
  doi: {{%
10\hspace{.1pt}\discretionary{.}{%
}{.}\hspace{.4pt}1155\discretionary{/}{%
}{/}2022\discretionary{/}{%
}{/}5880959}}


\bibitem{munch_users_2017}
E.~Munch.
\newblock A {User}'s {Guide} to {Topological} {Data} {Analysis}.
\newblock {\em Journal of Learning Analytics}, 4(2), July 2017. doi: {{%
10\hspace{.1pt}\discretionary{.}{%
}{.}\hspace{.4pt}18608\discretionary{/}{%
}{/}jla\hspace{.1pt}\discretionary{.}{%
}{.}\hspace{.4pt}2017\hspace{.1pt}\discretionary{.}{%
}{.}\hspace{.4pt}42\hspace{.1pt}\discretionary{.}{%
}{.}\hspace{.4pt}6}}


\bibitem{muszynski_topological_2019}
G.~Muszynski, K.~Kashinath, V.~Kurlin, M.~Wehner, and {Prabhat}.
\newblock Topological data analysis and machine learning for recognizing
  atmospheric river patterns in large climate datasets.
\newblock {\em Geoscientific Model Development}, 12(2):613--628, Feb. 2019.
  doi: {{%
10\hspace{.1pt}\discretionary{.}{%
}{.}\hspace{.4pt}5194\discretionary{/}{%
}{/}gmd\discretionary{%
}{-}{-}12\discretionary{%
}{-}{-}613\discretionary{%
}{-}{-}2019}}


\bibitem{nanni_deep_2025}
L.~Nanni, S.~Brahnam, M.~Ruta, D.~Fabris, M.~Boscolo~Bacheto, and T.~Milanello.
\newblock Deep {Ensembling} of {Multiband} {Images} for {Earth} {Remote}
  {Sensing} and {Foramnifera} {Data}.
\newblock {\em Sensors}, 25(7):2231, Apr. 2025. doi: {{%
10\hspace{.1pt}\discretionary{.}{%
}{.}\hspace{.4pt}3390\discretionary{/}{%
}{/}s25072231}}


\bibitem{neumann_-domain_2019}
M.~Neumann, A.~S. Pinto, X.~Zhai, and N.~Houlsby.
\newblock In-domain representation learning for remote sensing, Nov. 2019.
\newblock arXiv:1911.06721 [cs]. doi: {{%
10\hspace{.1pt}\discretionary{.}{%
}{.}\hspace{.4pt}48550\discretionary{/}{%
}{/}arXiv\hspace{.1pt}\discretionary{.}{%
}{.}\hspace{.4pt}1911\hspace{.1pt}\discretionary{.}{%
}{.}\hspace{.4pt}06721}}


\bibitem{ofori-boateng_application_2021}
D.~Ofori-Boateng, H.~Lee, K.~M. Gorski, M.~J. Garay, and Y.~R. Gel.
\newblock Application of {Topological} {Data} {Analysis} to
  {Multi}-{Resolution} {Matching} of {Aerosol} {Optical} {Depth} {Maps}.
\newblock {\em Frontiers in Environmental Science}, 9:684716, June 2021. doi:
  {{%
10\hspace{.1pt}\discretionary{.}{%
}{.}\hspace{.4pt}3389\discretionary{/}{%
}{/}fenvs\hspace{.1pt}\discretionary{.}{%
}{.}\hspace{.4pt}2021\hspace{.1pt}\discretionary{.}{%
}{.}\hspace{.4pt}684716}}


\bibitem{rolnick_tackling_2023}
D.~Rolnick, P.~L. Donti, L.~H. Kaack, K.~Kochanski, A.~Lacoste, K.~Sankaran,
  A.~S. Ross, N.~Milojevic-Dupont, N.~Jaques, A.~Waldman-Brown, A.~S. Luccioni,
  T.~Maharaj, E.~D. Sherwin, S.~K. Mukkavilli, K.~P. Kording, C.~P. Gomes,
  A.~Y. Ng, D.~Hassabis, J.~C. Platt, F.~Creutzig, J.~Chayes, and Y.~Bengio.
\newblock Tackling {Climate} {Change} with {Machine} {Learning}.
\newblock {\em ACM Computing Surveys}, 55(2):1--96, Feb. 2023. doi: {{%
10\hspace{.1pt}\discretionary{.}{%
}{.}\hspace{.4pt}1145\discretionary{/}{%
}{/}3485128}}


\bibitem{sena_topological_2021}
C.~{\'A}.~P. Sena, J.~A.~R. Da~Paix{\~a}o, and J.~R. D.~A. Fran{\c{c}}a.
\newblock A {Topological} {Data} {Analysis} approach for retrieving {Local}
  {Climate} {Zones} patterns in satellite data.
\newblock {\em Environmental Challenges}, 5:100359, Dec. 2021. doi: {{%
10\hspace{.1pt}\discretionary{.}{%
}{.}\hspace{.4pt}1016\discretionary{/}{%
}{/}j\hspace{.1pt}\discretionary{.}{%
}{.}\hspace{.4pt}envc\hspace{.1pt}\discretionary{.}{%
}{.}\hspace{.4pt}2021\hspace{.1pt}\discretionary{.}{%
}{.}\hspace{.4pt}100359}}


\bibitem{singh_topological_2023}
Y.~Singh, C.~M. Farrelly, Q.~A. Hathaway, T.~Leiner, J.~Jagtap, G.~E. Carlsson,
  and B.~J. Erickson.
\newblock Topological data analysis in medical imaging: current state of the
  art.
\newblock {\em Insights into Imaging}, 14(1):58, Apr. 2023. doi: {{%
10\hspace{.1pt}\discretionary{.}{%
}{.}\hspace{.4pt}1186\discretionary{/}{%
}{/}s13244\discretionary{%
}{-}{-}023\discretionary{%
}{-}{-}01413\discretionary{%
}{-}{-}w}}


\bibitem{tauzin_giotto-tda_2021}
G.~Tauzin, U.~Lupo, L.~Tunstall, J.~B. P{\'e}rez, M.~Caorsi, W.~Reise,
  A.~Medina-Mardones, A.~Dassatti, and K.~Hess.
\newblock giotto-tda: {A} {Topological} {Data} {Analysis} {Toolkit} for
  {Machine} {Learning} and {Data} {Exploration}, Mar. 2021.
\newblock arXiv:2004.02551 [cs]. doi: {{%
10\hspace{.1pt}\discretionary{.}{%
}{.}\hspace{.4pt}48550\discretionary{/}{%
}{/}arXiv\hspace{.1pt}\discretionary{.}{%
}{.}\hspace{.4pt}2004\hspace{.1pt}\discretionary{.}{%
}{.}\hspace{.4pt}02551}}


\bibitem{ver_hoef_primer_2023}
L.~Ver~Hoef, H.~Adams, E.~J. King, and I.~Ebert-Uphoff.
\newblock A {Primer} on {Topological} {Data} {Analysis} to {Support} {Image}
  {Analysis} {Tasks} in {Environmental} {Science}.
\newblock {\em Artificial Intelligence for the Earth Systems}, 2(1):e220039,
  Jan. 2023. doi: {{%
10\hspace{.1pt}\discretionary{.}{%
}{.}\hspace{.4pt}1175\discretionary{/}{%
}{/}AIES\discretionary{%
}{-}{-}D\discretionary{%
}{-}{-}22\discretionary{%
}{-}{-}0039\hspace{.1pt}\discretionary{.}{%
}{.}\hspace{.4pt}1}}


\bibitem{wang_mtp_2024}
D.~Wang, J.~Zhang, M.~Xu, L.~Liu, D.~Wang, E.~Gao, C.~Han, H.~Guo, B.~Du,
  D.~Tao, and L.~Zhang.
\newblock {MTP}: {Advancing} {Remote} {Sensing} {Foundation} {Model} via
  {Multitask} {Pretraining}.
\newblock {\em IEEE Journal of Selected Topics in Applied Earth Observations
  and Remote Sensing}, 17:11632--11654, June 2024. doi: {{%
10\hspace{.1pt}\discretionary{.}{%
}{.}\hspace{.4pt}1109\discretionary{/}{%
}{/}JSTARS\hspace{.1pt}\discretionary{.}{%
}{.}\hspace{.4pt}2024\hspace{.1pt}\discretionary{.}{%
}{.}\hspace{.4pt}3408154}}


\bibitem{yuan_deep_2020}
Q.~Yuan, H.~Shen, T.~Li, Z.~Li, S.~Li, Y.~Jiang, H.~Xu, W.~Tan, Q.~Yang,
  J.~Wang, J.~Gao, and L.~Zhang.
\newblock Deep learning in environmental remote sensing: {Achievements} and
  challenges.
\newblock {\em Remote Sensing of Environment}, 241:111716, May 2020. doi: {{%
10\hspace{.1pt}\discretionary{.}{%
}{.}\hspace{.4pt}1016\discretionary{/}{%
}{/}j\hspace{.1pt}\discretionary{.}{%
}{.}\hspace{.4pt}rse\hspace{.1pt}\discretionary{.}{%
}{.}\hspace{.4pt}2020\hspace{.1pt}\discretionary{.}{%
}{.}\hspace{.4pt}111716}}


\bibitem{zia_topological_2024}
A.~Zia, A.~Khamis, J.~Nichols, U.~B. Tayab, Z.~Hayder, V.~Rolland, E.~Stone,
  and L.~Petersson.
\newblock Topological deep learning: a review of an emerging paradigm.
\newblock {\em Artificial Intelligence Review}, 57(4):77, Feb. 2024. doi: {{%
10\hspace{.1pt}\discretionary{.}{%
}{.}\hspace{.4pt}1007\discretionary{/}{%
}{/}s10462\discretionary{%
}{-}{-}024\discretionary{%
}{-}{-}10710\discretionary{%
}{-}{-}9}}


\end{thebibliography}
\end{document}